\documentclass[sigconf]{acmart}
\AtBeginDocument{%
	}

\copyrightyear{2023}
\acmYear{2023}
\setcopyright{acmlicensed}\acmConference[MM '23]{Proceedings of the 31st ACM International Conference on Multimedia}{October 29-November 3, 2023}{Ottawa, ON, Canada}
\acmBooktitle{Proceedings of the 31st ACM International Conference on Multimedia (MM '23), October 29-November 3, 2023, Ottawa, ON, Canada}
\acmPrice{15.00}
\acmDOI{10.1145/3581783.3611816}
\acmISBN{979-8-4007-0108-5/23/10}

\acmSubmissionID{534}
\usepackage[american]{babel}
\usepackage{microtype}
\usepackage{balance}

\usepackage{ragged2e} 
\usepackage{booktabs,makecell, multirow, tabularx}
\usepackage{colortbl}
\usepackage{bbding}
\usepackage{bm}
\usepackage{algorithm}
\usepackage{algorithmic}

\newcommand{\M}{DSSN }

\newcommand{\argmax}{\mathop{\rm arg~max}\limits}

\begin{document}
	\title{Improving Semi-Supervised Semantic Segmentation with Dual-Level Siamese Structure Network}
	\author{Zhibo Tian}
	\email{tianzhb21@lzu.edu.cn}
	\affiliation{%
		\institution{School of Information Science and Engineering, \\
			Lanzhou University}
		\city{Lanzhou}
		\country{China}
	}

	\author{Xiaolin Zhang}
\email{solli.zhang@gmail.com}
\affiliation{%
	\institution{Independent Researcher}
	\city{Shenzhen}
	\country{China}
}

	\author{Peng Zhang}
\email{pengzhang_skd@sdust.edu.cn}
\affiliation{%
	\institution{College of Computer Science and Engineering, \\
		Shandong University of Science and Technology}
	\city{Qingdao}
	\country{China}
}
	
	\author{Kun Zhan}
	\authornote{Corresponding author.}
	\email{kzhan@lzu.edu.cn}	
	\affiliation{%
		\institution{School of Information Science and Engineering, \\
			Lanzhou University}
		\city{Lanzhou}
		\country{China}
	}
\renewcommand{\shortauthors}{Zhibo Tian, Xiaolin Zhang, Peng Zhang, \& Kun Zhan}
	
\newcommand{\xiaolin}{\textcolor[rgb]{0,0,1}}
\newcommand{\peng}{\textcolor[rgb]{1,0,0}}
\newcommand{\etal}{\textrm{et~al.\,}}
\newcommand{\eg}{\textrm{e.g.}}
\newcommand{\ie}{\textrm{i.e.}}
\newcommand{\wrt}{\textit{w.r.t.\,}}

\newcommand{\x}{\bm{x}}
\newcommand{\z}{\bm{z}}
\newcommand{\h}{\bm{h}}
\newcommand{\p}{\bm{y}}
\newcommand{\auglow}{{\rm AugL}}
\newcommand{\aughigh}{{\rm AugH}}
	
	\begin{abstract}
		Semi-supervised semantic segmentation (SSS) is an important task that utilizes both labeled and unlabeled data to reduce expenses on labeling training examples. However, the effectiveness of SSS algorithms is limited by the difficulty of fully exploiting the potential of unlabeled data. To address this, we propose a dual-level Siamese structure network (DSSN) for pixel-wise contrastive learning. By aligning positive pairs with a pixel-wise contrastive loss using strong augmented views in both low-level image space and high-level feature space, the proposed DSSN is designed to maximize the utilization of available unlabeled data. Additionally, we introduce a novel class-aware pseudo-label selection strategy for weak-to-strong supervision, which addresses the limitations of most existing methods that do not perform selection or apply a predefined threshold for all classes. Specifically, our strategy selects the top high-confidence prediction of the weak view for each class to generate pseudo labels that supervise the strong augmented views. This strategy is capable of taking into account the class imbalance and improving the performance of long-tailed classes. Our proposed method achieves state-of-the-art results on two datasets, PASCAL VOC 2012 and Cityscapes, outperforming other SSS algorithms by a significant margin. The source code is available at \url{https://github.com/kunzhan/DSSN}.
	\end{abstract}


	
	\maketitle

	\begin{figure}[h]
		\centering
		\includegraphics[width=0.9\linewidth]{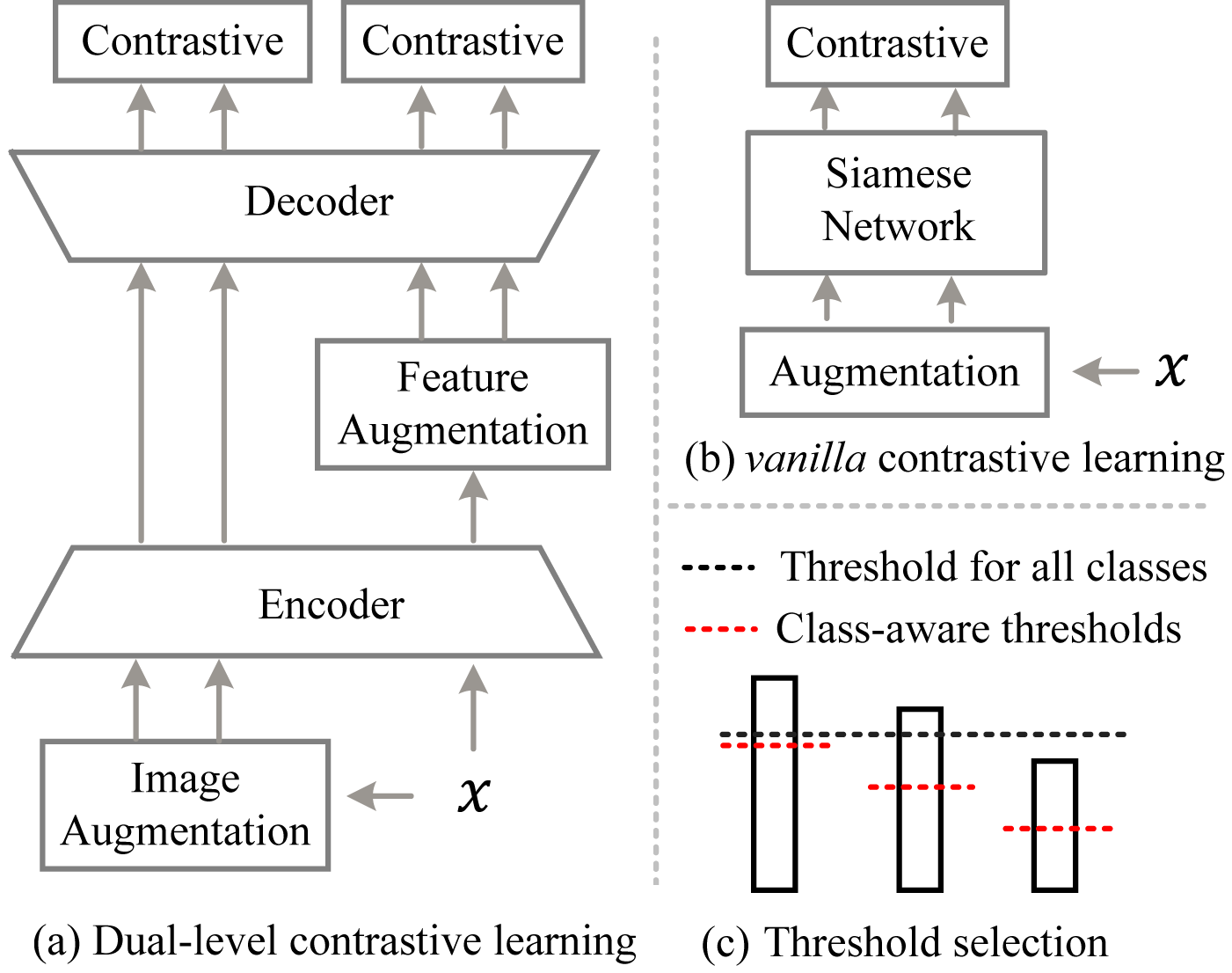}
		\caption{Illustration of the motivation. (a) demonstrates the proposed dual-level contrastive structure for exploiting the maximum potential of unlabelled samples. (b) depicts the structure of the \textit{vanilla} contrastive learning. (c) compares the threshold selection strategies of the proposed class-aware pseudo-label generation method and the classical approaches of utilizing a threshold for all classes.}
		\label{motivation}
	\end{figure}

	\section{Introduction}
	Deep learning methods for supervised segmentation have shown remarkable performance. However, they heavily rely on a large amount of annotated images, which is labor cost and time-consuming. Alternatively, semi-supervised semantic segmentation (SSS) offers a viable solution to address this fundamental weakness by exploiting the readily available unlabeled data to improve model performance.

	Existing semi-supervised learning methods typically use unlabeled samples in two ways: pseudo supervision~\cite{berthelot2019mixmatch,sohn2020fixmatch} and consistency regularization~\cite{laine2016temporal,tarvainen2017mean,xie2020unsupervised}.  
	Pseudo supervision is to generate pseudo labels for the unlabeled images and gradually incorporates them into the training process to supervise model learning. 
	For example, preliminary works~\cite{hung2018adversarial,mittal2019semi} in SSS tend to utilize the generative adversarial networks~\cite{creswell2018generative} as auxiliary supervision for unlabeled images.
	Consistency regularization promotes agreement among model predictions on unlabeled samples that are subjected to various perturbations, thus improving model generalization by ensuring that different views of the same unlabeled image are consistent. Modern SSS algorithms combine pseudo supervision and consistency regularization into a two-view network architecture, where one view generates pseudo labels to supervise the other view for prediction consistency. For instance, the intuition of CPS~\cite{chen2021semi} is that using one view generates pseudo labels of unlabeled images to expand the training set of the other view. PseudoSeg~\cite{zou2020pseudoseg} generates pseudo labels in a weak augmented view to supervise the other strong augmented view. PS-MT~\cite{liu2022perturbed} employs higher-confidence pseudo labels than CPS by averaging the predictions of two views. 
		To search for high-quality pseudo labels, CCT~\cite{ouali2020semi} employs a fixed threshold for all classes and pixels with confidence scores above the threshold to participate in network updates. In CCT~\cite{ouali2020semi}, it mainly uses consistency learning between one weak view and two strong augmented views of a high-level feature.
	
	However, many existing SSS algorithms do not fully exploit the potential of unlabelled data. To address this issue, we propose a Dual-level Siamese structure network (DSSN) to fully exploit feature diversities. In addition to the two strategies commonly used in most algorithms, we introduce a new variant of contrastive learning.
	Fig.~\ref{motivation}(b) illustrates a typical structure of the \textit{vanilla} contrastive learning, which excels at providing extraordinary generalization abilities for unlabeled samples~\cite{LeCun1467314,Hadsell2006}. Specifically, 
		the proposed DSSN simultaneously employs pixel-wise contrastive learning and two-level strong augmented views. Accordingly, contrastive objectives in terms of image-level and feature-level augmentations are introduced to guide the network training. Such structure guarantees fully exploiting the potential of unlabeled data.
	As shown in Fig.~\ref{motivation}(a), at the image level, two different views of unlabeled samples are obtained with different strong augmentations, and a pixel-wise contrastive objective is added to train DSSN using the corresponding predictions. At the feature level, high-level latent features from the encoder produce two strong augmented views and also conduct a contrastive loss. This DSSN design enables us to fully exploit the available unlabeled data.
	
	Given that most real-world datasets exhibit imbalanced or long-tailed label distributions~\cite{menonlong}, we propose a class-aware pseudo label generation (CPLG) strategy that selects class-specific high-confidence pseudo labels from weak views to supervise the strong views. Our CPLG strategy differs from previous approaches~\cite{french2019semi,ouali2020semi}, which apply a fixed threshold to all categories. By treating each class differently, our method aims to improve the performance of long-tailed categories. Without any selection, low-quality pseudo labels generated from the weak augmented view are used to supervise the strong augmented view, which could negatively affect the model training. Using a constant threshold for all classes may result in long-tailed classes being poorly trained, as their confidence may be lower than the threshold and thus not involved in training. Using a fixed threshold may also result in useful pseudo-labels being ignored in some classes that fall below the predefined threshold. For each class has pseudo labels, we select top high-confidence pixels in each class since most segments in an image are imbalances and also it is imbalances in the whole dataset.  A schematic illustrating this strategy is presented in Fig.~\ref{motivation}(c). This approach increases the contribution of long-tailed classes and addresses the learning difficulties of different classes.
	
	In summary, DSSN makes the following contributions:
	
	(1) DSSN offers a novel approach to leverage unlabeled data in training SSS models by utilizing dual-level pixel-wise contrastive learning. This approach is a valuable addition to the existing techniques of exploiting unlabeled data, such as pseudo-supervision and consistency regularization.
	
	(2) DSSN's design enables the maximal utilization of available unlabeled data. The dual-level structure is not only utilized in contrastive learning but also in weak-to-strong pseudo-supervision. 
	
	(3) We introduce a novel class-aware pseudo-label selection strategy for weak-to-strong supervision, known as CPLG. This strategy effectively improves the performance of long-tailed classes.
	\section{Related Work} 
SSS has two mainstream methods, pseudo supervision and consistency regularization. Preliminary works~\cite{hung2018adversarial,mittal2019semi} use the generative adversarial networks~\cite{creswell2018generative} to generate pseudo supervision. Specifically, consistency regularization methods encourage consistency prediction of unlabeled samples with various perturbation. The CutMix-Seg~\cite{french2019semi} approach incorporates the CutMix~\cite{yun2019cutmix} augmentation into semantic segmentation in order to supply consistency restrictions on unlabeled data and also revealed Cutout~\cite{devries2017improved} and CutMix~\cite{yun2019cutmix} are critical to the success of consistency regularization. Alternatively, CCT~\cite{ouali2020semi} proposes a feature-level perturbation and a cross-consistency training method that enforce consistency between the main decoder predictions and auxiliary decoders. By using two segmentation models with the same structure but different initialization, GCT~\cite{ke2020guided} conducts network perturbation and promotes consistency between the predictions from the two models. In the meantime, CPS~\cite{chen2021semi} constructs two parallel networks to provide cross-pseudo labels for one another. DMT~\cite{feng2022dmt} re-weights the loss on different regions based on the disagreement of two different initialized models. Self-training by pseudo labeling is a classic technique that dates back about a decade, taking the most likely class as a pseudo label and training models on unlabeled data is a common method for achieving minimum entropy. Concurrently  ST++~\cite{yang2022st++} also demonstrates that employing suitable data perturbations on unlabeled samples is really quite beneficial for self-training. Unimatch~\cite{unimatch}  explores the effectiveness of weak-to-strong supervision, leveraging dual strong augmentations.
	
 Contrastive learning is one of the alternative methods that stands out. RoCo~\cite{liu2021bootstrapping} and U$^2$PL~\cite{wang2022semi} use InfoNCE loss~\cite{oord2018representation} on the predicted logits, but they not use Siamese structure network as shown in Fig.~\ref{motivation}(b). DSSN obtains better performance than them, which can be seen in the experiment section. 
	
	\begin{figure*}[h]
\centering
\includegraphics[width=0.82\linewidth]{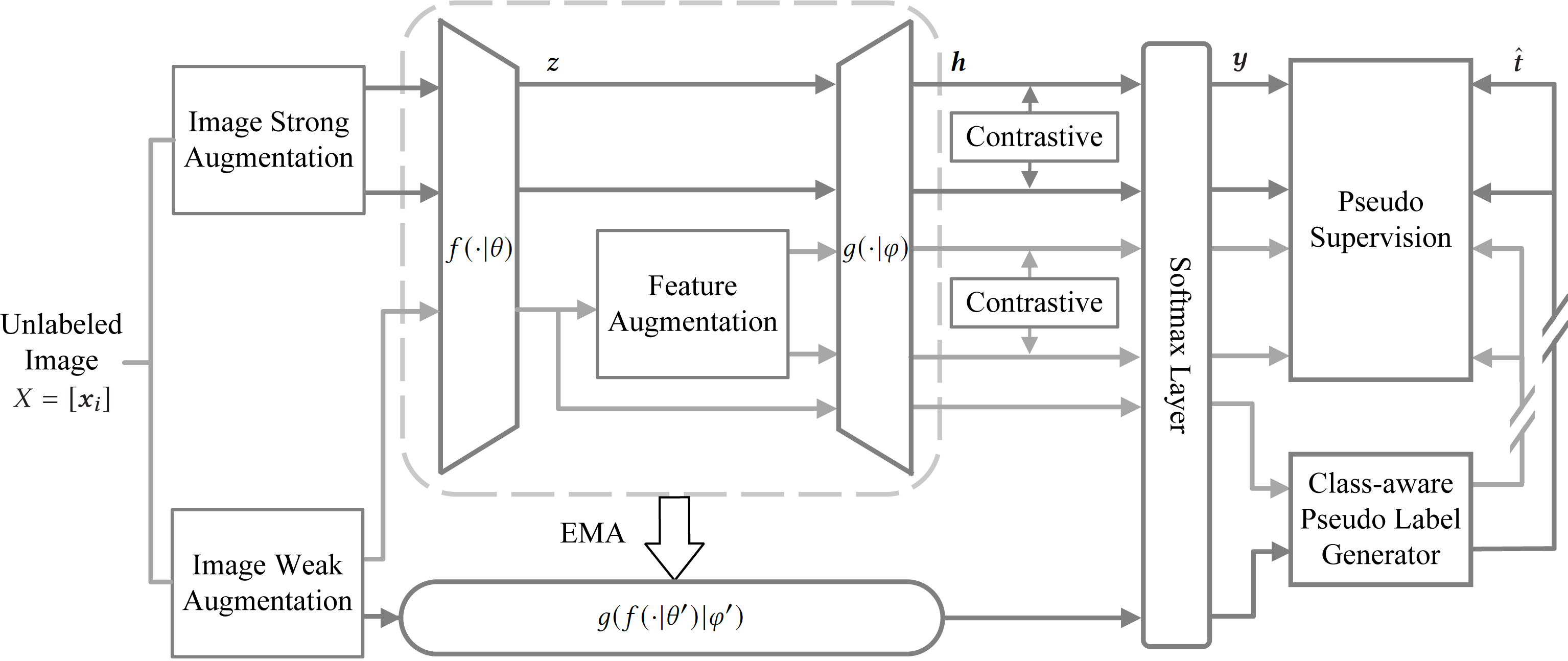}
\caption{The overview of \M. Dual-level contrastive learning and weak-to-strong pseudo supervision.}
\label{framework}
	\end{figure*}
	\section{Method} 
	\subsection{Preliminaries}
Following SSS works~\cite{chen2021semi,yang2022st++,liu2021bootstrapping}, we use both a small fraction of labeled data $\mathcal D_l=\{(X_i,\bm{T}_i)\}_{i=1}^M$ and a large fraction of unlabeled data $\mathcal D_u=\{X_i\}_{i=1+M}^{N+M}$\,. $X_i$ denotes an image, and $\bm{T_i}$ represents its ground-truth label if $X_i$ is a labeled image. 
	$N$ and $M$ indicate the number of labeled and unlabeled images, respectively, where $N \gg M$ in most cases. 
	To facilitate the calculation of loss functions, we represent each pixel in an image as a vector $\x$ since a pixel has values in different channels. Thus, in subsequent sections, we represent each pixel as a vector $\x$ with $\bm{t}$ as its one-hot ground-truth label. Given an image $X=[\x_i]$ with the size of $W \times H$ where $W$ and $H$ are the width and height, we denote the pixel by $\x_i, i \in \{1,...,W \times H\}$. The latent high-level feature $\z$ corresponding to $\x$ is obtained by an encoder $f(\x|\theta)$ where $\theta$ is the learnable parameters of the encoder. We yield the predicted logits $\h$ by feeding the latent representations $\z$ into a decoder $g(\z|\varphi)$ where $\varphi$ is the learnable parameters of the decoder. Finally, a softmax layer is added to obtain the ultimate probability for each class,~\ie, $\p={\rm softmax}(\h)$.

	Given a labeled image, we use a supervised cross-entropy loss,
	\begin{align}
		\mathcal L_{\rm sup}=-\sum_i\sum_{j\in \mathcal C} t_{ij}\log y_{ij}\label{loss_sup}
	\end{align}
	where $\mathcal C=\{1,\ldots,C\}$ and $C$ is the total number of classes.
	For a unlabeled image,  a simple way to generate their pseudo labels $\hat{\bm{t}}_{i}$  is to apply a one-hot operation to the predictions,~\ie, $\p_{i}$.
	For the $i$-th pixel of an unlabeled image, we represent the predicted probability of the $i$-th pixel belonging to the $j$-th class as $y_{ij}$\,. 
	Specifically, we use the following operation to generate pseudo labels:
	\begin{align}
		c &= \argmax_{j\in \mathcal C}(y_{ij}),
		\label{eq_conf}\\
		\hat{t}_{ij}&=
		\begin{cases}
			1, &{\rm if}\,j=c\\
			0, &{\rm otherwise}
		\end{cases}
		\label{eq_pseduo}
	\end{align}
	where $c$ denotes the maximal probability within the class $j\in \mathcal C$, the $\hat{\bm{t}}_i=[\hat{t}_{ij}]$ is the one-hot pseudo label.
	\subsection{Dual-Level Contrastive Learning}
	To fully exploit the potential of available unlabeled data, 
	we propose to use DSSN for extracting pixel-wise contrastive positive pairs in different abstraction levels. The low-level image is subjected to two-view strong augmentations,
	\begin{align}
		\x_{i}^{ls1}&=\auglow_s(\x_i),\\
		\x_{i}^{ls2}&= \auglow_s(\x_i)
	\end{align}
	where $\x_{i}^{ls1}$ denotes the strong augmented low-level pixel in the first view. The output, $\auglow_s(\cdot)$, is random. $\auglow_s(\cdot)$ generates varying outputs using the same input to augment the data diversity. This increases the diversity, resulting in an improvement in the robustness and generalization ability of the training model.
	
We use two-view augmented images to obtain its decoded logits,
	\begin{eqnarray}
		\h_i^{ls1}&=&g(f(\x_i^{ls1}|\theta)|\varphi)\,.\\
		\h_i^{ls2}&=&g(f(\x_i^{ls2}|\theta)|\varphi)\,.
	\end{eqnarray}

	Analogous to~\cite{hjelm2018learning}, we apply the contrastive objective,~\ie, $\mathcal L_{\rm cl}$ to pairwise pixels for learning better representations:
	\begin{align}
		\mathcal L_{\rm cl}
		=&- \frac{1}{|\mathcal P|}\sum_{(i,i)\in\mathcal P} \log d(\h_i^{ls1},\h_i^{ls2})\notag\\
		&-\frac{1}{|\mathcal N|}\sum_{{(i,j)\in\mathcal N}}  \log\bigl(1-d(\h_i^{ls1},\h_j^{ls2})\bigr)\label{loss_cl}
	\end{align}
	where $d(\cdot,\cdot)$ is a similarity score of a pair of logits. $\h_i^{ls1}$ and $\h_i^{ls2}$ are belong to positive pairs $(i,i)\in\mathcal P$ while $\h_i^{ls1}$ and  $\h_j^{ls2}$ are negative pairs $(i,j)\in\mathcal N, \forall\,i\neq j$. We use $\mathcal P$ and $\mathcal N$ to denote the sets of positive and negative pairs, respectively.
	
	Inspired by BYOL~\cite{grill2020bootstrap}, we only use the positive pairs. The similarity $d(\cdot,\cdot)$ of positive logits is defined by a Gaussian function, 
	\begin{align}\label{eq-gaussian}
		d(\h^{ls1}_i,\h_i^{ls2})=\exp\Bigl(-{\bigl\|\h^{ls1}_i-\h_i^{ls2}\bigr\|^2_2}\Bigr)\,.
	\end{align}
The similarity defined by Eq.~\eqref{eq-gaussian} implies the similarity is 1 if the pairwise logits are the same while it tends to 0 if their distance is far from each other. From a different perspective, the error $\|\h^{ls1}_i-\h_i^{ls2}\bigr\|^2_2$ of two-view logits is governed by the Gaussian distribution due to the central limit theorem~\cite{Walker1969PML}, so we also obtain Eq.~\eqref{eq-gaussian}.
	
	Substituting Eq.~\eqref{eq-gaussian} into Eq.~\eqref{loss_cl} obtains the following loss.
	\begin{align}
		\mathcal L^{ls}_{\rm cl}
		=\frac1{W\times H}\sum_i\bigl\|\h^{ls1}_i-\h_i^{ls2}\bigr\|^2_2
		\label{loss_cll}
	\end{align}
	where we only use pixel-wise positive pairs. 
	
	For the high-level feature contrastive learning, we obtain the high-level latent feature with the encoder,
	\begin{align}
		\z_i^{hw} = f(\auglow_w(\x_i)|\theta)
	\end{align}
	where $\auglow_w(\cdot)$ is a weak augmentation for the low-level pixel. The high-level feature is subjected to two-view strong augmentations,
	\begin{align}
		\z_{i}^{hs1}&= \aughigh_s(\z_i^{hw}),\\
		\z_{i}^{hs2}&= \aughigh_s(\z_i^{hw})
	\end{align}
	
	We use the two-view augmented features to obtain its decoded logits, $\h_i^{hs1}=g(\z_{i}^{hs1}|\varphi)$ and $\h_i^{hs2}=g(\z_{i}^{hs12}|\varphi)$\,. Then, we use them to construct the contrastive loss,
	\begin{align}
		\mathcal L^{hs}_{\rm cl}
		=\frac1{W\times H}\sum_i\bigl\|\h^{hs1}_i-\h_i^{hs2}\bigr\|^2_2\,.
		\label{loss_clh}
	\end{align}
	\subsection{Weak-to-Strong Pseudo Supervision}
	To leverage the four predictions generated by a strongly augmented image, we feed the corresponding weakly augmented image into DSSN. Next, we use the prediction of the weak view to generate its pseudo label and supervise the four strong views. Given our dual-level structure, weak-to-strong pseudo supervision is also performed in both levels. Specifically, we use the pseudo labels of the weak view, denoted as $\hat{\bm{t}}$, to supervise the predictions of the strong views, denoted as $\p$,.
	
	The weak pseudo supervisions are obtained by
	\begin{align}
\p^{lw}&={\rm softmax}(g(f(\x|\theta')|\varphi'))\\
\p^{hw}&={\rm softmax}(g(\z^{hw}|\varphi))
	\end{align}
	where $(\theta',\varphi')$ of the teacher are updated from the student $(\theta,\varphi)$ by the exponential moving average (EMA)
	\begin{align}
		(\theta',\varphi') = \alpha(\theta',\varphi') + (1-\alpha)(\theta,\varphi) 
		\label{EMA}
	\end{align}
	where $\alpha$ is a momentum parameter, with $\alpha\in[0,1]$.
	
	The pseudo labels $\hat{\bm{t}}^{lw}$ and $\hat{\bm{t}}^{hw}$ of $\p^{lw}$ and $\p^{hw}$ are calculated by using Eqs.~\eqref{eq_conf} and \eqref{eq_pseduo}, respectively. 
	
	The output probability of the strong augmented views, $\p_{i}^{ls1}$, $\p_{i}^{ls2}$, $\p_{i}^{hs1}$, and $\p_{i}^{hs2}$, are attained by the softmax layer. 
	
	The weak-to-strong pseudo-supervision loss functions are
	\begin{align}
		\mathcal L^{(l)}_{\rm w2s}&=-\sum_{i}\sum_j m_{ij}^{lw}\left(\hat{t}_{ij}^{lw}\log y_{ij}^{ls1}+\hat{t}_{ij}^{lw}\log y_{ij}^{ls2}\right)\label{loss_w2s_l}\\
		\mathcal L^{(h)}_{\rm w2s}&=-\sum_{i}\sum_j m_{ij}^{hw}\left(\hat{t}_{ij}^{hw}\log y_{ij}^{hs1}+\hat{t}_{ij}^{hw}\log y_{ij}^{hs2}\right)\label{loss_w2s_h}
	\end{align}
	where $m_{ij}$ is a class-wise binary mask to select the pixel with high-confidence score and we show how to obtain it in the next section.
	\subsection{Class-aware pseudo-label generation}\label{sbs_cplg}
	As shown in Fig.~\ref{motivation}(c), we show the class-aware pseudo-label generation (CPLG). For the $i$-th pixel, it has different probabilities belonging to different classes. $y_{ij}$ denotes the probability of the $i$-th pixel belonging to the $j$-th class. We observe all pixels in the same class, \ie, in the same channel of network output. 
	
	First, we find the pixel class-wisely that has the largest probability in the $j$-th class,
	\begin{equation}
		y_j^{\max}=\max_{i}(y_{ij})\,,\forall\,j\in\mathcal C\,.
		\label{eq_maxclass}
	\end{equation}
	
	Second, we establish a class-wise threshold $\tau_j$ by multiplying the maximum probability by $r$\%. Pixels exceeding this class-wise threshold are selected. Additionally, we restrict the maximum probability by $\tau_{\rm low}$ and exclude pixels with a low maximum probability since they indicate lower prediction confidence. Thus, the class-wise threshold $\tau_j$ is determined by
	\begin{equation}
		\tau_j =
		\begin{cases}
			y_j^{\max}\cdot r\%,&{\rm if}\,y_j^{\max}>\tau_{\rm low}\\
			y_j^{\max}, &{\rm otherwise}
		\end{cases}\label{eq_threshold}
	\end{equation}
	where the ratio $r$ and the low bound $\tau_{\rm low}$ are parameters.
	
	Third, we select pixels in each class by $\tau_j$, \ie, pixels exceeding $\tau_j$ are selected:
	\begin{equation}
		m_{ij}=
		\begin{cases}
			1, &{\rm if}\,y_{ij}>\tau_j\\
			0, &{\rm otherwise}\,.
		\end{cases}
		\label{eq_mask}
	\end{equation}
	
	The generation of the pseudo label is straightforward by using Eqs.~\eqref{eq_conf} and \eqref{eq_pseduo}. The refined class-aware pseudo labels are attained by multiplying them, \ie, $m_{ij}\hat{t}_{ij}$, as used in Eqs.~\eqref{loss_w2s_l} and \eqref{loss_w2s_h}. Our CPLG strategy considers the learning status and difficulties of different classes by adjusting thresholds for each class. As a result, we select useful pixels with low thresholds for training, which enhances the accuracy of challenging classes. 
	
\subsection{Overall Algorithm}
Fig.~\ref{framework} illustrates how we combine two distinct learning strategies for the unlabeled images: contrastive learning and weak-to-strong pseudo supervision. 
	
In this section, we present the DSSN algorithm, which is illustrated in Algorithm~\ref{algorithm}. It takes a small fraction of labeled data and a large fraction of unlabeled data as input to train the model. The supervised loss between the model prediction on labeled data and the ground truth is computed using Eq.~\eqref{loss_sup}. Subsequently, the low-level and high-level contrastive learning losses are calculated using Eqs.~\eqref{loss_cll} and \eqref{loss_clh}, respectively. We then compute the weak-to-strong pseudo-supervision loss using Eqs.~\eqref{loss_w2s_l} and \eqref{loss_w2s_h}. The overall loss term is formulated as follows:
	\begin{equation}\label{all_loss}
		\mathcal L=\mathcal L_{\rm sup}+\gamma_1 \Bigl(\mathcal L_{\rm cl}^{ls}+\mathcal L_{\rm cl}^{hs}\Bigr)+\gamma_2 \Bigl(\mathcal L^{(l)}_{\rm w2s}+\mathcal L^{(h)}_{\rm w2s}\Bigr),
	\end{equation}
	where $\gamma_1$ and $\gamma_2$ are the trade-off weight. Finally, we update the student model and the teacher model by using the error back-propagation algorithm and EMA, respectively.
	
	\begin{algorithm}[!ht]
		\caption{The DSSN algorithm.}\label{algorithm}
		\begin{algorithmic}[1]
			\STATE \textbf{Input}: $\mathcal D=\{\mathcal D_u, \mathcal D_l\}$, and batch size $b$.
			\STATE \textbf{Output}: $(\theta',\varphi')$\,.
			\STATE \textbf{Initialization}: $epoch=0$, $epoch_{\max}$, and $(\theta,\varphi)$\,.
			\WHILE{$epoch\leq {epoch_{\max}}$}
			\FOR{mini-batch samples in $\mathcal D$}
			\STATE Feed the samples into DSSN for forward propagation\,;
			\STATE Update $\mathcal L_{\rm sup}$ by Eq.~\eqref{loss_sup}\,;
			\STATE Update $\mathcal L_{\rm cl}^{ls}$ and $\mathcal L_{\rm cl}^{hs}$ by Eqs.~\eqref{loss_cll} and \eqref{loss_clh}\,;
			\STATE Update $\mathcal L^{(l)}_{\rm w2s}$ and $\mathcal L^{(h)}_{\rm w2s}$ by Eqs.~\eqref{loss_w2s_l} and \eqref{loss_w2s_h}\,;
			\STATE Update $\mathcal L$ by Eq.~\eqref{all_loss}\,;
			\STATE Update $(\theta,\varphi)$ by back propagation of $\sum_b\mathcal L$\,;
			\STATE Update $(\theta',\varphi')$ by Eq.~\eqref{EMA}\,;
			\STATE $epoch = epoch + 1$\,;
			\ENDFOR
			\ENDWHILE
		\end{algorithmic}
	\end{algorithm}
	
	\section{Experiments}
	In this section, we first present the details of the experiments. Second, we compare the proposed \M method to the recent state-of-the-art (SOTA) approaches to the SSS task. Third, we conduct extensive ablation experiments to demonstrate the effectiveness and robustness of the proposed method.
	\subsection{Experimental setup}
	\textbf{Datasets.} We evaluate the proposed method on two classical semantic segmentation datasets,~\ie, PASCAL VOC 2012~\cite{everingham2015pascal} and Cityscapes~\cite{cordts2016cityscapes}.
	In particular, PASCAL VOC 2012~\cite{everingham2015pascal} has 20 classes of objects and 1 class of background. The standard training, validation and test sets consist of 1,464, 1449 and 1,456 images, respectively. Following the previous work~\cite{yang2022st++,chen2021semi, ke2020guided}, we also use augmented set SBD~\cite{hariharan2011semantic} (9,118 images) and original training set (1,464 images) as our full training set (10,582 images). The labels from the SBD~\cite{hariharan2011semantic} dataset are noise-prone and of low quality.  
	Cityscapes~\cite{cordts2016cityscapes} has 19 semantic classes and is mostly intended for understanding urban scenes. 
	It consists of 500 validation images, 1,525 test images, and 2,975 training images.
	All of the images have well-annotated masks.
	For a fair comparison with the benchmarks, we follow the partition procedure of CPS~\cite{chen2021semi}.
	Specifically, the training set is divided into two partitions by randomly sampling 1/2, 1/4, 1/8, and 1/16 of the total set as the labeled samples and the remaining images as the unlabeled for the blended set.
	
	\textbf{Implementation details.} 
	Following the previous benchmarks CPS~\cite{chen2021semi}, we adopt DeepLab~v3+~\cite{chen2018encoder} based on ResNet~\cite{he2016deep} as the segmentation network for a fair comparison.
	The backbone~\ie, ResNet, is initialized with the weights pre-trained on ImageNet~\cite{deng2009imagenet}.
	The segmentation heads are randomly initialized. 
	During training, each mini-batch contains eight labeled and eight unlabeled images. The stochastic gradient descent (SGD) optimizer is used, and the initial learning rates are set to 0.002 and 0.005 for the PASCAL VOC 2012 and Cityscapes, respectively. In accordance with other works~\cite{chen2021semi,ouali2020semi}, we employ the following polynomial to decrease the learning rate while training: $(1 - {{epoch}}/{{epoch_{\max}}})^{0.9}$. The model is trained for 100 epochs on PASCAL VOC 2012 and 240 epochs for Cityscapes. For \textit{weak augmentations}, we adopt the same operation as ST++\cite{yang2022st++}, where the training images are random flipping and resizing (between 0.5 and 2.0 times), followed by a random crop operation to the resolutions of  513 $\times$ 513 and 801 $\times$ 801 for the two datasets, respectively. 
	We employ several \textit{strong augmentation}, including random color-jitter, grayscale, Gaussian blur, etc. For \textit{strong feature augmentation}, we apply a random dropout of 50\% on features from the encoder. The unsupervised trade-off weights $\gamma_1$ and $\gamma_2$ are set as 0.01 and 0.25. In CPLG, $r$ is set to 96\% and $\tau_{\rm low}$ is 0.92, respectively.
	
	Additionally, we also apply CutMix~\cite{yun2019cutmix} data augmentation to the student model images. The EMA smoothing factor $\alpha$ is set as 0.996. 
	We follow U$^2$PL~\cite{wang2022semi}, the supervised loss is cross-entropy on PASCAL, and for Cityscapes the cross-entropy loss is replaced by the online hard example mining loss. 
	
	\textbf{Evaluation.}  We use the mean of Intersection-over-Union(mIoU) for the validation set to evaluate the segmentation performance for both datasets. 
	Following the previous works~\cite{chen2021semi,yang2022st++}, 
	we employ the sliding evaluation to examine the efficacy of our model on the validation images from Cityscapes with a resolution of 1024×2048. 
	
	\begin{table}[!ht]
		\centering
		\setlength{\tabcolsep}{2pt} 
		\caption{ Comparison with SOTAs with ResNet-101. Labeled images are from the original high-quality original training set of PASCAL VOC 2012. 
		}
		\begin{tabular}{lccccc}
			\toprule
			Method 									 & 1/16(92) & 1/8(183)   & 1/4(366)    & 1/2(732)    & Full(1464) \\
			\midrule
			Baseline                                & 44.10   & 52.30      & 61.80       & 66.70       & 72.90  \\
			CutMix-Seg~\cite{french2019semi} 		 & 52.16   & 63.47      & 69.46       & 73.73       & 76.54 \\
			
			PseudoSeg~\cite{zou2020pseudoseg}		 & 57.60   & 65.50      & 69.14       & 72.41       & 73.23 \\
			PC$^2$Seg~\cite{zhong2021pixel} 		 & 57.00   & 66.28      & 69.78       & 73.05       & 74.15 \\
			CPS~\cite{chen2021semi} 				 & 64.07   & 67.42      & 71.71       & 75.88       & - \\
			ReCo~\cite{liu2021bootstrapping}         & 64.78   & 72.02      & 73.14       & 74.69       & - \\
			PS-MT~\cite{liu2022perturbed} 			 & 65.80   & 69.58      & 76.57       & 78.42       & 80.01 \\
			ST++~\cite{yang2022st++} 				 & 65.20   & 71.00      & 74.60       & 77.30       & 79.10  \\
			U$^2$PL~\cite{wang2022semi} 			 & 67.98   & 69.15      & 73.66       & 76.16       & 79.49 \\
			PCR~\cite{xu2022semi} 					 & 70.06   & 74.71      & 77.16       & 78.49       & 80.65 \\
			GTA-Seg~\cite{jin2022semi}				 & 70.02   & 73.16      & 75.57       & 78.37       & 80.47 \\
			Unimatch~\cite{unimatch} 				 & 75.20   & 77.20      & 78.80       & 79.90       & 81.20 \\
			\hline
			\textbf{DSSN}  	                  &{75.24}        &{76.75}            &{78.68}   & {80.61}       & {81.18} \\
			\bottomrule
		\end{tabular}%
		\label{tb_voc_1446}
	\end{table}%
	\subsection{Comparison to SOTA Methods}
	\begin{table*}[!tp]
		\centering
		\caption{Comparison with the state-of-the-art methods on blended PASCAL VOC 2012 under different partition protocols. 
		}
		\begingroup 
		\setlength{\tabcolsep}{6pt} 
		\renewcommand{\arraystretch}{1.1} 
		\begin{tabular}{l|c|c|c|c|c|c|c|c}
			\toprule[1pt]
			\multirow{2}[2]{*}{Method}       &                  \multicolumn{4}{c|}{ResNet-50}                   &                \multicolumn{4}{c}{ResNet-101}                 \\ \cline{2-9}
			&   1/16 (662)   &   1/8 (1323)   &   1/4 (2646)   &   1/2 (5291)   &   1/16 (662)   &   1/8 (1323)   &   1/4 (2646)   & 1/2 (5291) \\ \hline
			Baseline                         &61.20           &67.30           &70.80           & 74.75           &65.6            & 70.40           &72.80           & 76.65      \\ \hline
			MT~\cite{tarvainen2017mean}      &     66.77      &     70.78      &     73.22      &     75.41      &     70.59      &     73.20      &     76.62      &   77.61    \\
			CutMix-Seg~\cite{french2019semi} &     68.90      &     70.70      &     72.46      &     74.49      &     72.56      &     72.69      &     74.25      &   75.89    \\
			CCT~\cite{ouali2020semi}       &     65.22      &     70.87      &     73.43      &     74.75      &     67.94      &     73.00      &     76.17      &   77.56    \\
			GCT~\cite{ke2020guided}          &     64.05      &     70.47      &     73.45      &     75.20      &     69.77      &     73.30      &     75.25      &   77.14    \\
			CPS~\cite{chen2021semi}           &     71.98      &     73.67      &     74.90      &     76.15      &     74.48      &     76.44      &     77.68      &   78.64    \\
			ST++~\cite{yang2022st++}          &     72.60      &     74.40      &     75.40      &       -        &     74.50      &     76.30      &     76.60      &     -      \\
			U$^2$PL~\cite{wang2022semi}       &     72.00      &     75.10      &      76.20     &       -        &     74.43      &     77.60      &     78.70      &   -    \\
			PS-MT~\cite{liu2022perturbed}     &     72.83      &     75.70      &     76.43      &     77.88      &     75.50      &     78.20      &     78.72      &   79.76\    \\
			Unimatch~\cite{unimatch}          &     75.80      &     76.90      &     76.80      &       -        &     78.10      &     78.40      &     79.20      &     -      \\
			\hline 
			\textbf{DSSN}                             & {76.74} & {77.81} & {78.27} & {78.32} & {78.50} & {79.58} & {79.45} &  {79.96}\\ 
			\bottomrule[1pt]
		\end{tabular}
		\endgroup
		\label{res_voc}
	\end{table*}
	
\begin{table*}[!tp]
\centering
\caption{Comparison with state-of-the-art on Cityscapes, $*$ means the reproduced results in U$^2$PL~\cite{wang2022semi}.}
\begingroup 
\setlength{\tabcolsep}{7pt} 
\renewcommand{\arraystretch}{1.1} 
\begin{tabular}{l|c|c|c|c|c|c|c|c}
\toprule[1pt]
\multirow{2}[2]{*}{Method}      &                  \multicolumn{4}{c|}{ResNet-50}                   &                \multicolumn{4}{c}{ResNet-101}                \\ \cline{2-9}
&   1/16 (186)   &   1/8 (372)    &   1/4 (744)    &   1/2 (1488)   &   1/16 (186)   &   1/8 (372)    & 1/4 (744) &   1/2 (1488)   \\ \hline
Baseline                        &63.30           &70.20           &73.10           &76.60            &66.30           &72.80           &75.00       &78.00           \\ \hline
MT~\cite{tarvainen2017mean}      &     66.14      &     72.03      &     74.47      &     77.43      &     68.08      &     73.71      &   76.53   &     78.59      \\
CutMix-Seg~\cite{french2019semi} &       -        &       -        &       -        &       -        &     72.13      &     75.83      &   77.24   &     78.95      \\
CCT~\cite{ouali2020semi}       &     66.35      &     72.46      &     75.68      &     76.78      &     69.64      &     74.48      &   76.35   &     78.29      \\
GCT~\cite{ke2020guided}          &     65.81      &     71.33      &     75.30      &     77.09      &     66.90      &     72.96      &   76.45   &     78.58      \\
CPS~$^*$~\cite{chen2021semi}        &       -        &       -        &       -        &       -        &     69.78      &     74.31      &   74.58   &     76.81      \\
ST++~\cite{yang2022st++}         &       -        &     72.70      &      73.8      &       -        &       -        &       -        &     -     &       -        \\
U$^2$PL~\cite{wang2022semi}      &       69.03      &    73.02      &       76.31     &       78.64    &     70.30      &     74.37      &   76.47   &  79.05      \\
PS-MT~\cite{liu2022perturbed}    &       -        &     75.76      &     76.92      &     77.64      &       -        &     76.89      &   77.60   &     79.09      \\
Unimatch~\cite{unimatch}         &     75.00      &     76.80      &     77.50      &     78.60      &     76.60      &     77.90      &   79.20   &     79.50      \\
\hline
\textbf{DSSN}                            & 75.41 & 77.31 & 78.05 & 78.73& 76.52 & {78.18} &  {78.62}           & {79.58} \\ \bottomrule[1pt]
\end{tabular}%
\endgroup
\label{res_city}
\end{table*}
	
	\begin{figure}[htbp]
		\centering
		\includegraphics[width=\linewidth]{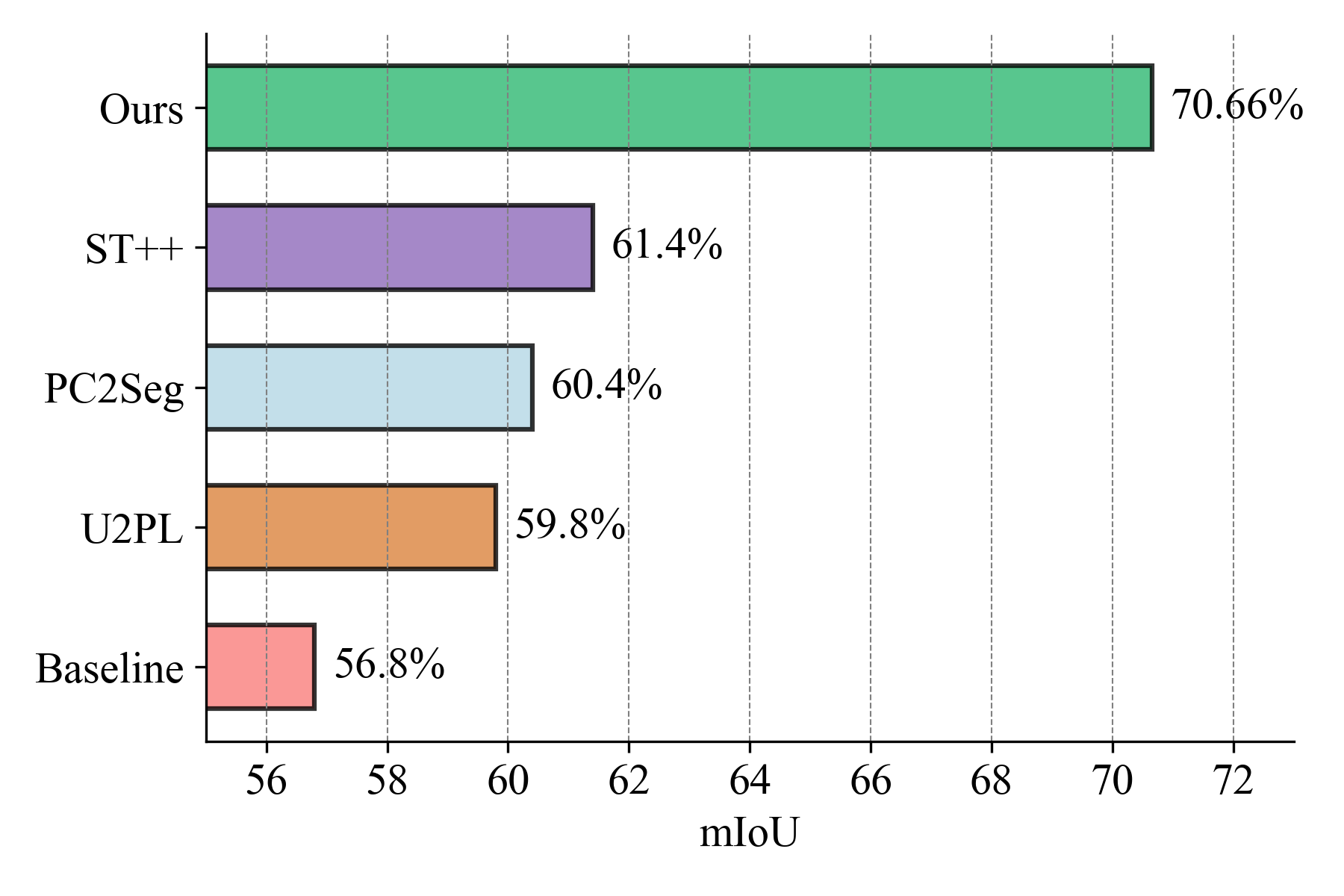}
		\caption{The proposed DSSN method effectively utilizes unlabeled images, as demonstrated by its performance on the Cityscapes dataset with a 1/30 data split and ResNet-50. Compared to SOTAs, DSSN outperforms them significantly.}\label{1_30_res50}
	\end{figure}
	
	To demonstrate the superiority of our proposed~\M~method, we conduct a comparison with the current state-of-the-art methods across various settings. All results are reported on the validation set for both PASCAL VOC and Cityscapes datasets. Additionally, we present the corresponding baseline at the top of the table, representing the results of purely supervised learning trained on the same labeled data. To ensure a fair comparison, all methods employed the DeepLab~v3+ architecture.
	
	\textbf{PASCAL VOC 2012.} We report results of our experiments on the PASCAL VOC 2012 validation dataset in Tables~\ref{tb_voc_1446} and \ref{res_voc}, where we evaluate the mean Intersection over Union (mIoU) for different proportions of labeled samples. Additionally, we present the corresponding baseline at the top of the table, representing the results of purely supervised learning trained on the same labeled data.
	
	Table~\ref{tb_voc_1446} presents results on the classic PASCAL VOC 2012 dataset. It shows our method significantly outperforms current state-of-the-art methods. When employing ResNet-101 as the backbone, DSSN attains a 5.18\% performance gain on the 1/16(92) split which surpass the performance obtain by the (1/3)183 data split in the prior study. Even with more labeled data, the performance differences become less evident; however, the proposed method still demonstrates performance improvements of 2.21\% with 1/2 fine annotations over the previous SOTAs.
	
	Table~\ref{res_voc} illustrates the results on blender PASCAL VOC 2012 Dataset. Our method shows significant improvement on the 1/16, 1/8, 1/4, and 1/2 splits with ResNet-50, compared to the baseline, with improvements of 15.51\%, 10.1\%, 6.73\%, and 3.57\%, respectively. Similarly, with ResNet-101, our method achieves improvements of 12.9\%, 9.18\%, 6.65\%, and 3.01\% under the same partitions. Especially, our method shows significant improvements when the ratio of labeled data becomes smaller, such as under 1/8 or 1/16 partition protocols. In particular, when the labeled data is extremely limited,e.g., on the 1/16 partitions, our method achieves remarkable increases of 15.51\% and 12.9\% compared to the baseline with ResNet-50 and ResNet-101 as the backbone networks, respectively. Furthermore, our method demonstrates a considerable improvement over the previous state-of-the-art PS-MT~\cite{liu2022perturbed}, achieving a margin of 3.88\% with ResNet-50 as the backbone, and 1.7\% under the 1/8 partition protocol. 
	
	\textbf{Cityscapes.}
	In Table~\ref{res_city}, we can see that our method consistently outperforms the supervised baseline by a significant margin, achieving improvements of 12.11\%, 7.11\%, 4.95\%, and 2.13\% with ResNet-50 under 1/16, 1/8, 1/4, and 1/2 partition protocols, respectively. Similarly, with ResNet-101, our method shows improvements of 10.22\%, 5.38\%, 3.62\%, and 1.58\% under 1/16, 1/8, 1/4, and 1/2 partition protocols, respectively. Furthermore, our method outperforms all other state-of-the-art methods across various settings. Specifically, under 1/8, 1/4, and 1/2 partitions, DSSN achieves a 1.55\%, 1.13\%, and 1.09\% improvement over the previous state-of-the-art PS-MT~\cite{liu2022perturbed} using ResNet-50, and a 1.29\%, 1.02\%, and 0.49\% improvement using ResNet-101, respectively. 
	
	We evaluate DSSN using ResNet-50 on a 1/30 data split, which contained only 100 labeled images. As illustrated in Fig.~\ref{1_30_res50}, DSSN outperforms the current state-of-the-art significantly. This result indicates that our method effectively utilizes the unlabeled data through contrastive learning and the class-aware pseudo-label selection strategy (CPLG). Besides, although ReCo~\cite{liu2021bootstrapping} and U$^2$PL\cite{wang2022semi} try to  construct positive and negative pairs to use contrastive learning, the result shows our DSSN outperform them significantly.
	
	\begin{table}[!tp]
		\centering
		\caption{Ablation of contrastive learning and CPLG.}\label{tb_cc}
		\begin{tabular*}{0.42\textwidth}{@{\extracolsep{\fill}\,}ccc}
			\toprule
			$\mathcal L_{\rm cl}^{ls}+\mathcal L_{\rm cl}^{hs}$ & CPLG & mIoU      \\ 
			\hline
			&&\\
			\XSolidBrush   &     \XSolidBrush     &     76.12     \\
			\XSolidBrush    &     \Checkmark     &     78.33      \\
			\Checkmark    &     \XSolidBrush     &     78.70      \\
			\Checkmark    &      \Checkmark      & {79.58} \\ \hline
		\end{tabular*}
	\end{table}
	\begin{table}[!tp]
		\centering
		\caption{Ablation of low- and high-level contrastive learning.}\label{tb_abhl}
		\begin{tabular*}{0.42\textwidth}{@{\extracolsep{\fill}\,}ccc}
			\toprule
			$\mathcal L_{\rm cl}^{ls}$  & $\mathcal L_{\rm cl}^{hs}$ & mIoU \\ \hline
			&&\\
			\XSolidBrush   &     \XSolidBrush             &     78.33     \\
			\XSolidBrush   &     \Checkmark               &     78.90      \\
			\Checkmark     &     \XSolidBrush             &     79.19      \\
			\Checkmark     &      \Checkmark              & {79.58} \\ \hline
		\end{tabular*} 
	\end{table}

	\begin{figure}
	\centering
	\includegraphics[width=0.9\linewidth]{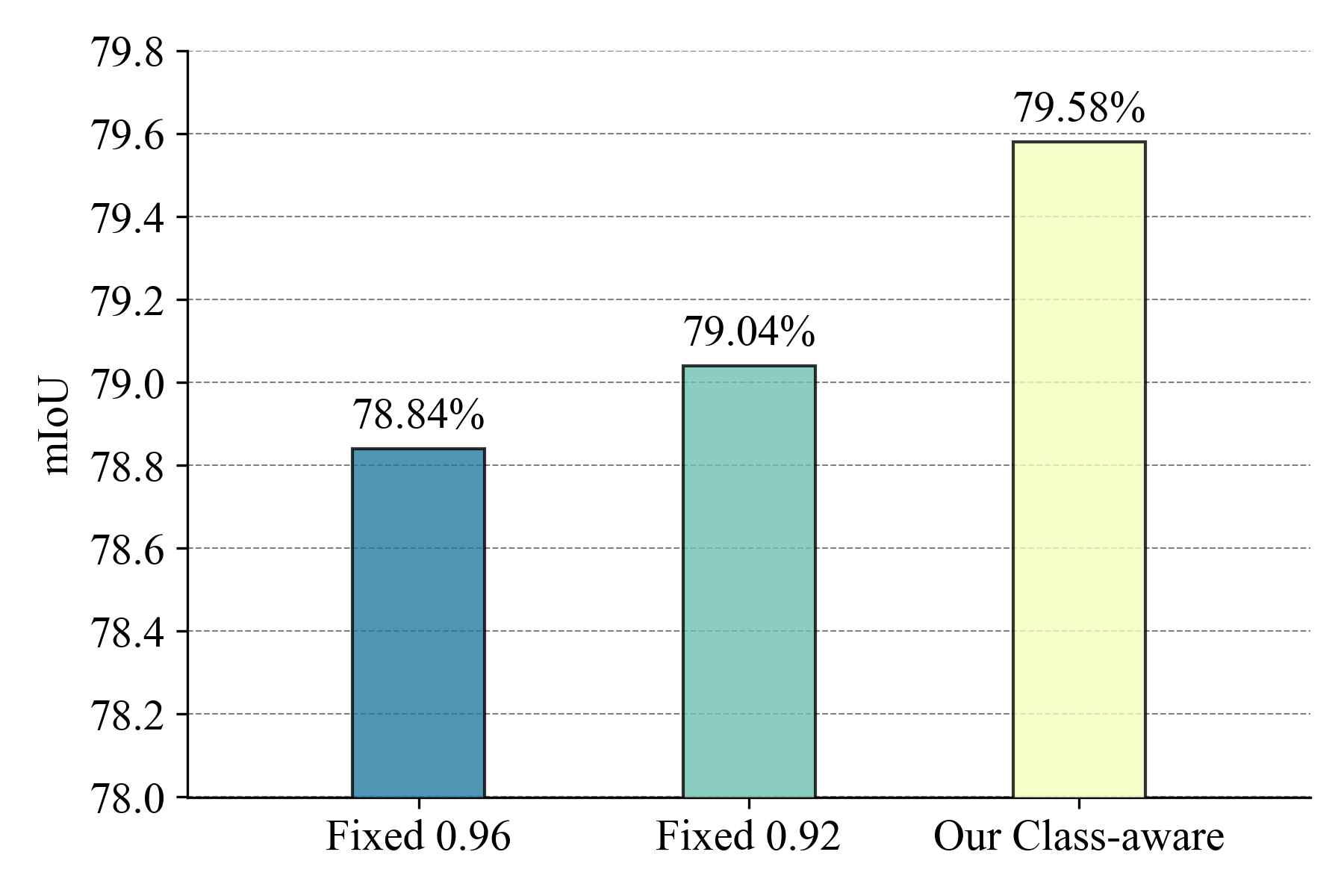}
	\caption{Comparsion CPLG to the fixed threshold.}
	\label{visual_threshold}
\end{figure}
\begin{figure}
	\centering
	\includegraphics[width=\linewidth]{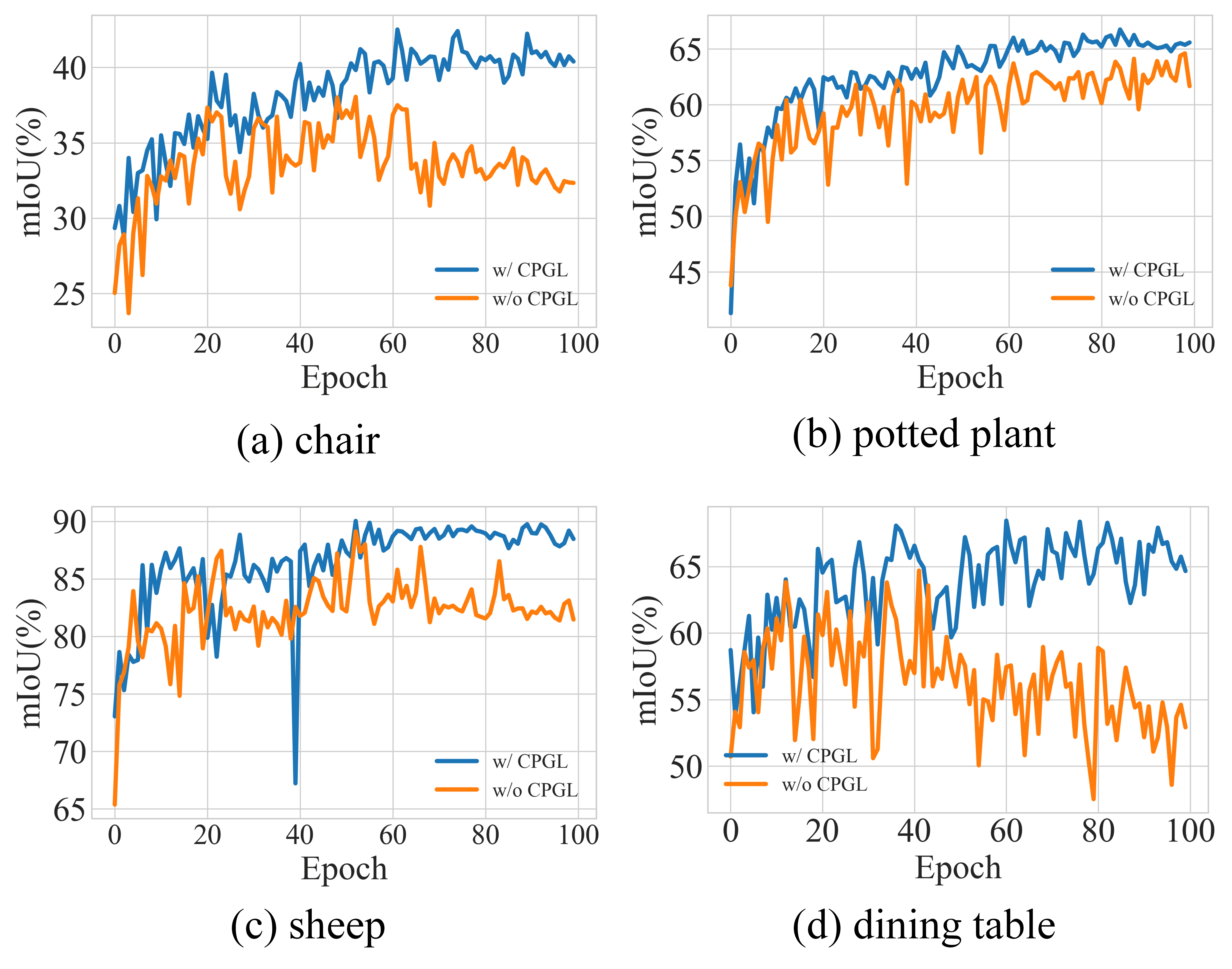}
	\caption{The mIoU of four long-tailed classes.}
	\Description{}
	\label{visual_classes_all}
\end{figure}
	Upon comparing performance on classic PASCAL VOC 2012 and blended training set, we observe that the quality of labeled samples is important. For example, DSSN achieves an exceptional performance of 80.61\% by utilizing only 732 high-quality labels. However, even with significantly more labels (5291) from the blended dataset, a comparable score of 80.61\% cannot be achieved.
	\subsection{Ablation Studies}
	In this subsection, we discuss the contribution of each component to our framework using ResNet-101 and a 1/8 labeled ratio on PASCAL VOC 2012 dataset.
	
	\textbf{Effectiveness of the DSSN components.} We conduct a step-by-step ablation study of each component to comprehensively assess their effectiveness. Table~\ref{tb_cc} presents the results of our study. Without our proposed dual-Level contrastive learning and CPLG, applying a plain consistency method yields an accuracy of 76.12\%. However, employing dual-level contrastive learning leads to an accuracy of 78.33\%, while the proposed CPLG results in 78.70\%. Combining both dual-level contrastive learning and CPLG produces the highest accuracy of 79.58\%, demonstrating the effectiveness of each component in the proposed DSSN method.

		\begin{figure}[!t]
		\centering
		\includegraphics[width=0.83\linewidth]{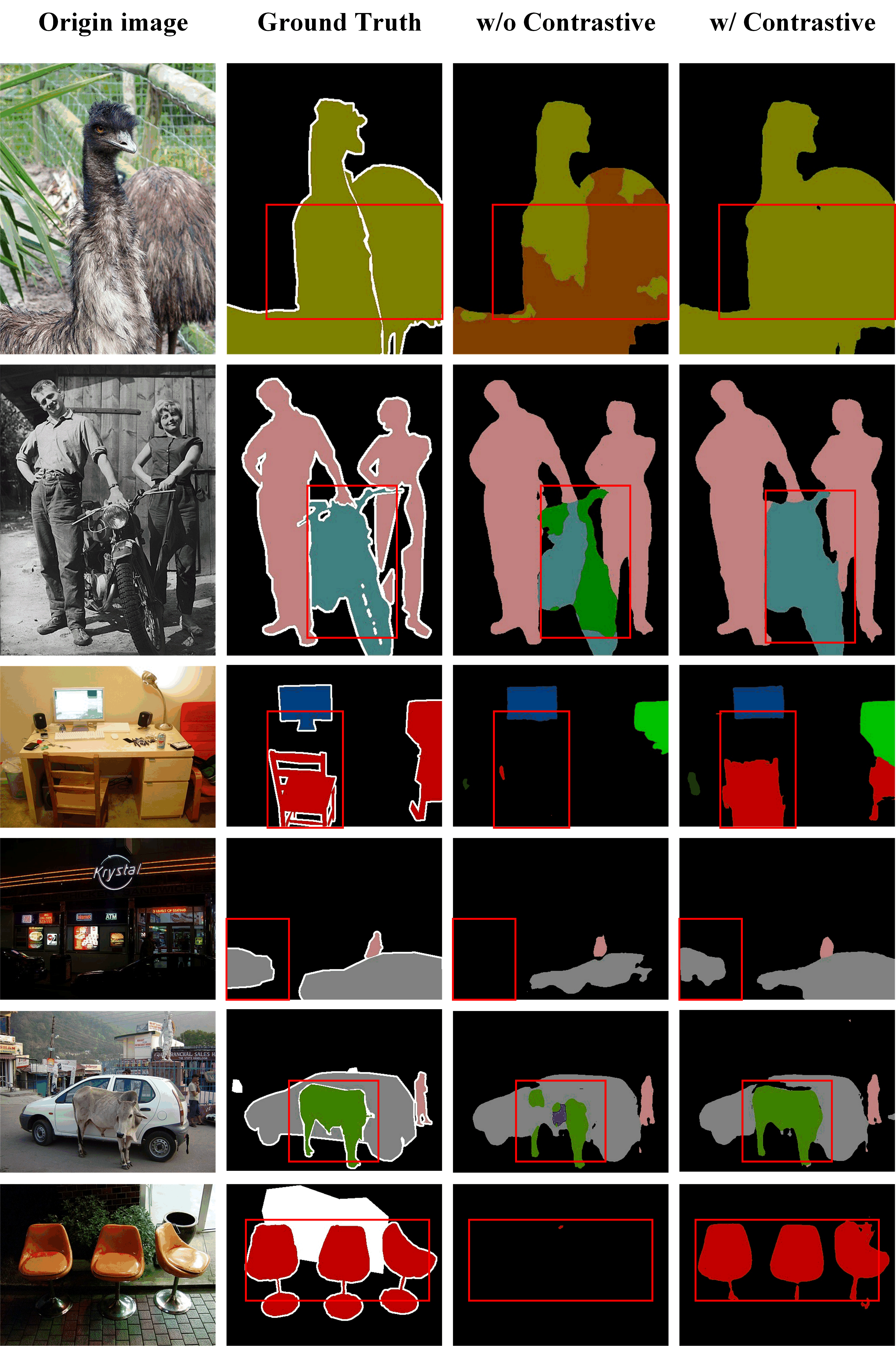}
		\caption{Visualization on PASCAL VOC 2012. Columns from left to right denote the input images, the ground-truth, DSSN without/with contrastive learning, respectively.}
		\label{visual_contr}
	\end{figure}

	\textbf{Effectiveness of contrastive Learning.}
	In our study, we incorporate both low-level and high-level contrastive learning in our dual-level contrastive learning approach. Table~\ref{tb_abhl} presents the results of our study. Without the use of both low-level contrastive and high-level contrastive, the accuracy was 78.33\%. Using low-level contrastive alone results in a 0.57\% improvement, while using high-level contrastive alone improves the accuracy by 0.86\%. Notably, using both low-level and high-level contrastive further improves the accuracy by 1.25\%, which shows the efficacy of our method.

\begin{figure}[!ht]
	\centering
	\includegraphics[width=0.95\linewidth]{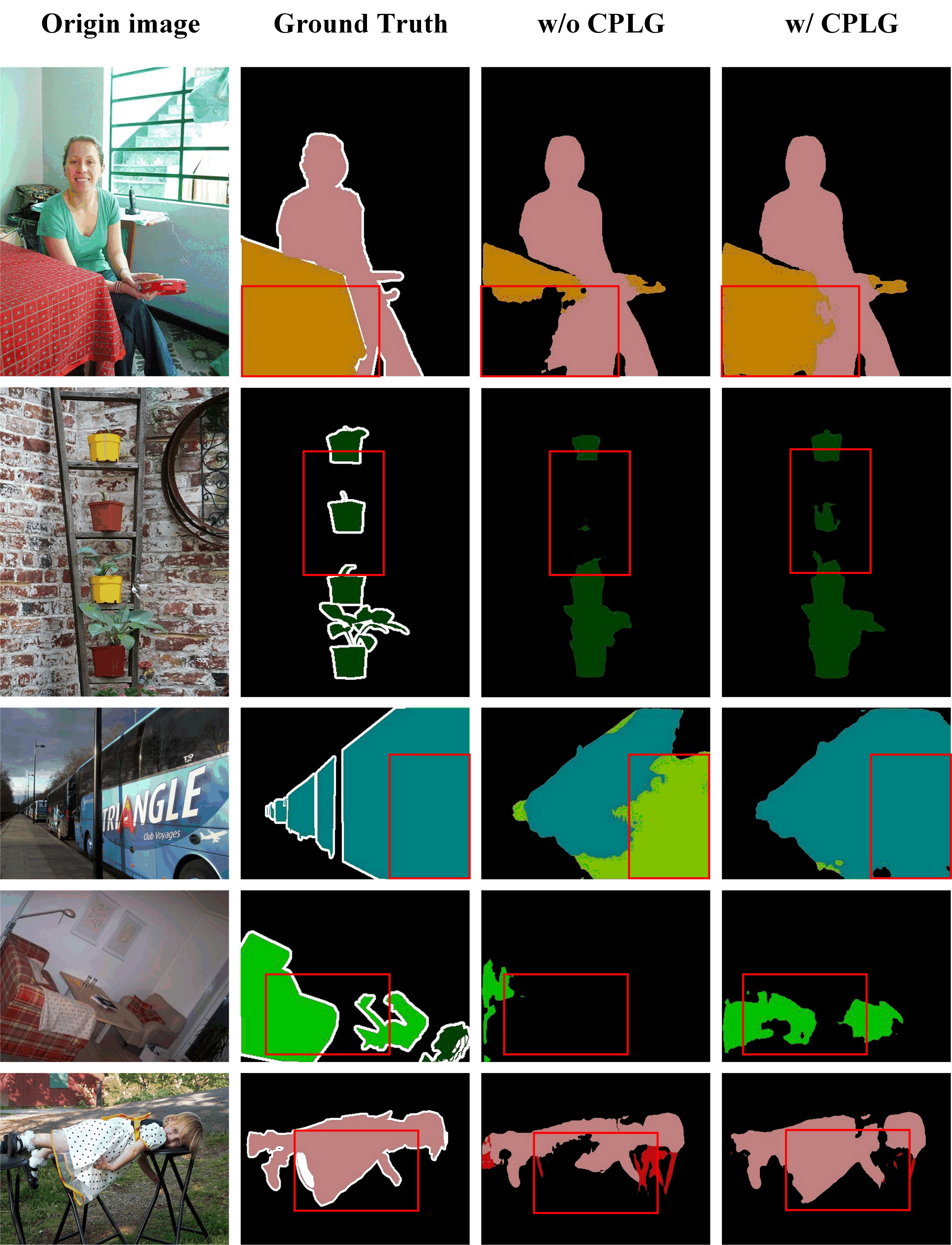}
	\caption{Visualization on PASCAL VOC 2012, from left to right, we show the raw images, the ground-truth, DSSN without/with CPLG, respectively.}
	\Description{}
	\label{visual_cplg}
\end{figure}
	
	\textbf{Effectiveness of CPLG.}
	As discussed in \S\ref{sbs_cplg}, the CPLG strategy considers difficulties of different classes and long-tailed classes, instead of using a fixed threshold during the pseudo-label generation. To test our method against a fixed threshold, we conduct experiments using a fixed threshold. Fig.~\ref{visual_threshold} shows that our strategy outperforms using a fixed threshold of 0.96 and 0.92 since we set $r$ to 0.96 and $\tau_{\rm low}$ to 0.92 in CPLG. This finding further highlights the effectiveness of our proposed DSSN method. We chose these specific thresholds because, following our experiments, we establish 0.92 as the lowest threshold and used 0.96 as the factor for the maximum probability value. Additionally, Fig.~\ref{visual_classes_all} presents mIoU values of classes with long tails and those that are hard to learn during training, which demonstrates the effectiveness of CPLG strategy.

	\textbf{Qualitative Results.} In Figs.~\ref{visual_contr} and \ref{visual_cplg}, we present the qualitative results of our study on the PASCAL VOC 2012 validation set. DSSN is based on the DeepLab~v3+ with ResNet-101 network and a 1/8 ratio. The integration of contrastive learning into our method improve the performance of our model for contour and ambiguous regions, while also enhancing the accuracy of some scenarios, as illustrated in Fig.~\ref{visual_contr}. Furthermore, our proposed CPLG achieved substantial precision in certain classes that are typically challenging to learn, as illustrated in Fig.~\ref{visual_cplg}.
	
	\section{Conclusion}
	In this paper, we introduce DSSN, a novel method that utilizes pixel-wise contrastive learning to address the SSS problem. DSSN is equipped with a dual-level structure that can effectively leverage unlabeled data. In DSSN, both contrastive learning and weak-to-strong consistency learning are utilized to maximize the utilization of available unlabeled data. Furthermore, we propose a class-aware pseudo label selection strategy that generates high-quality pseudo labels and significantly improves performance on long-tailed classes without incurring additional computation. DSSN achieves state-of-the-art performance on two benchmarks, and the effectiveness of our proposed novelties is confirmed by the ablation study.
\begin{acks}
	This work was supported by the National Natural Science Foundation of China under the Grant No. 62176108, Natural Science Foundation of Qinghai Province of China under No. 2022-ZJ-929, Fundamental Research Funds for the Central Universities under Nos. lzujbky-2021-ct09 and lzujbky-2022-ct06, Natural Science Foundation of Shandong Province of China, No. ZR2021QF017, and Supercomputing Center of Lanzhou University.
\end{acks}
	\bibliographystyle{ACM-Reference-Format}
	\balance
	\bibliography{egbib}
\end{document}